\documentclass[conference]{IEEEtran}
\IEEEoverridecommandlockouts
\IEEEaftertitletext{\vspace{-25pt}}
\usepackage{cite}
\usepackage{amsmath,amssymb,amsfonts}
\usepackage{algorithmic}
\usepackage{graphicx}
\usepackage{textcomp}
\usepackage{xcolor}

\allowdisplaybreaks
\usepackage{booktabs}
\usepackage[export]{adjustbox}
\usepackage{tcolorbox}
\usepackage{tikz}
\usepackage{ifthen}
\usetikzlibrary{fit}
\usetikzlibrary{positioning}
\usetikzlibrary{calc}
\usepackage{hyperref}
\usepackage{xurl}

\def\BibTeX{{\rm B\kern-.05em{\sc i\kern-.025em b}\kern-.08em
    T\kern-.1667em\lower.7ex\hbox{E}\kern-.125emX}}

\usepackage{fancyhdr}
\begin{document}
\bstctlcite{IEEEexample:BSTcontrol}

\title{A Computational Implementation of a Goal-Directed Theory of Affect
}
\author{\IEEEauthorblockN{1\textsuperscript{st} Bernhard Hilpert*}
\thanks{*Both authors contributed equally to this work}
\IEEEauthorblockA{\textit{Leiden Institute of Advanced Computer Science} \\
\textit{Leiden University}\\
b.hilpert@liacs.leidenuniv.nl}
\and
\IEEEauthorblockN{2\textsuperscript{nd} Tamás Szűcs*}
\IEEEauthorblockA{\textit{Research Group of Quantitative Psychology and Individual Differences} \\
\textit{KU Leuven}\\
tamas.szucs@kuleuven.be}
\and
\IEEEauthorblockN{3\textsuperscript{rd} Joost Broekens}
\IEEEauthorblockA{\textit{Leiden Institute of Advanced Computer Science} \\
\textit{Leiden University}\\
joost.broekens@gmail.com}
\and
\IEEEauthorblockN{4\textsuperscript{th} Agnes Moors}
\IEEEauthorblockA{\textit{Research Group of Quantitative Psychology and Individual Differences} \\
\textit{KU Leuven}\\
agnes.moors@kuleuven.be}
}

\maketitle
\thispagestyle{fancy}

\begin{abstract}
Computational modeling of emotion has long faced a tension between descriptive, "snapshot-based" appraisal models and granular, signal-driven architectures that often lack appropriate psychological grounding. This paper addresses this gap by presenting the first high-fidelity computational implementation of the Goal-Directed Theory (GDT) of affect. In this framework, affect is not a post-hoc label but a functional byproduct emerging from the continuous interplay between discrepancy detection and action selection within an agent’s internal processing cycles. We evaluate the model through a series of principled simulations (Dice/Corridor tasks) designed to isolate affective signatures and 
dynamics during multi-step goal pursuit. Results demonstrate that complex affective profiles, like an anticipatory "lift" and a failure "crash", emerge naturally from simple interactions between goal-discrepancy and action-selection expectancies without requiring additional dedicated modules. By ensuring every computational component maps directly to 
components of the psychological theory, this work establishes a transparent, testable framework that enables a continuous "simulation-empiry" research loop. Our work contributes to moving the field beyond "black-box" heuristics toward a granular, mechanistic understanding of affect, integrated into the core of agent behavior.
\end{abstract}

\vspace{-7pt}

\vspace{-12pt}
\section{Introduction}
Affect is the positive or negative aspect of subjective experience that we associate with moods, emotions, and other mental states \cite{kuppens_dynamic_2012, wundt_introduction_2013, russellCoreAffectPsychological2003a, schiller2024human}. We conceptualize affect as a bipolar dimension, characterized by a direction 
(positive/negative) and an intensity 
(distance from the neutral midpoint). 
While many approaches to explaining the generation of affect exist (for a comparative overview, see \cite{moors_demystifying_2022}), a computational implementation requires a specific level of mechanistic granularity in order to reflect the underlying theory faithfully. 
From a computational perspective, formalizing, e.g., appraisal-theoretical predictions requires specifying the mapping between specific appraisal patterns and specific affective outcomes, which are not always described with the required mechanistic granularity \cite{gratch2004domain,meuleman2013nonlinear, yeo2024associations}. 
By contrast, the goal-directed theory (GDT) of behavior and affect \cite{moors_demystifying_2022, moors2026emotions} describes a generative mechanism that derives affective outputs from inputs according to a set of underlying principles and processing cycles that specify the computation of affect from stimulus–goal discrepancies and expected utilities.
The GDT thus offers a promising parsimonious account that understands affect as arising within multiple goal-directed cycles designed to generate behavior, described with 
 a sufficient level of granularity in its mechanistic explanations making it a prime candidate for implementation in a computational framework.

Here, we present the first computational implementation of the GDT. We describe the model and the results of a simulation study demonstrating a number of features of the GDT computational model. 
This paper has two goals: 
1) to provide a high-fidelity computational implementation of the GDT, translating the psychological theory into a simulatable computational model
and 2) to simulate affective profiles that are comparable to human affect patterns in future experimental work, enabling an experimental loop between computational modeling and empirical validation.

\section{Background}

\subsection{The Goal-Directed Theory}
\begin{figure}[ht]
    \centering
    \includegraphics[width=0.9\linewidth, frame]{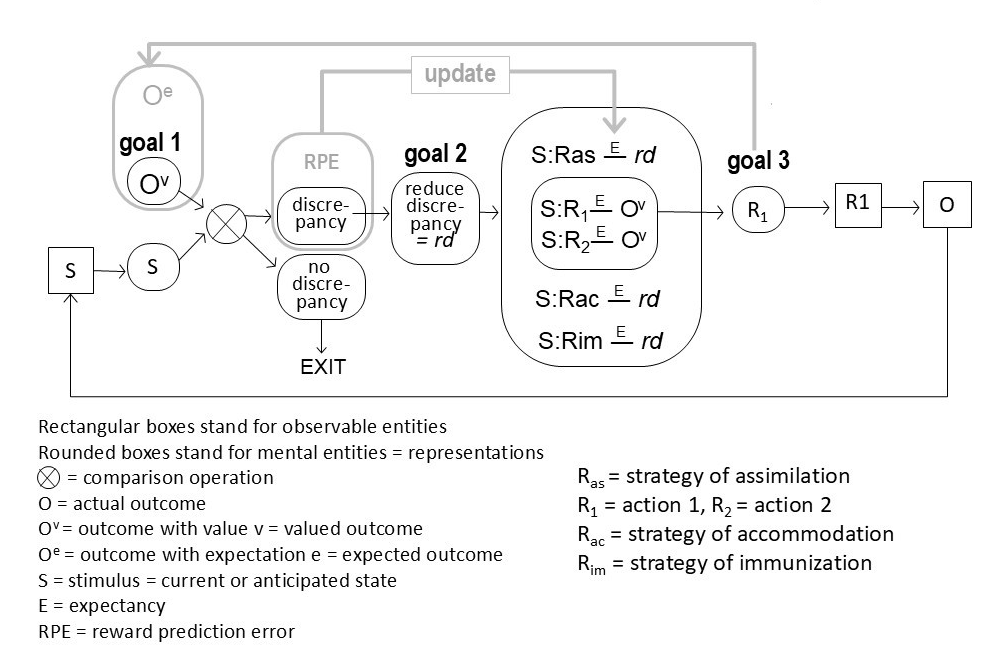}
    \vspace{-5pt}
    \caption{The goal-directed cycle}
    \label{fig:GDT}
    \vspace{-8pt}
\end{figure}

While traditional appraisal theories of emotion count as stimulus-evaluation theories, the GDT counts as a response-evaluation theory \cite{moors_demystifying_2022}. In appraisal theories, affect and other emotion components (action tendencies and physiological and motor responses) are predicted by appraisal patterns (i.e., values on appraisal dimensions such as goal congruence, goal relevance, expectedness, control, and agency, \cite{moors2013appraisal, scherer2001appraisal}). 
In the GDT, by contrast, affect and other emotion components are predicted by stimulus–goal discrepancies (corresponding to the appraisal dimensions of goal congruence and goal relevance) and by the expected utilities of available behavioral options \cite{moors2026emotions}.
A goal-directed cycle in the GDT consists of two phases: a stimulus-goal discrepancy detection phase and an action selection phase (Fig. \ref{fig:GDT}). The cycle starts with an actual or anticipated stimulus representation being compared to a first goal (i.e., representation of a valued outcome, $[O_1^V]$). If there is no discrepancy, the cycle stops. If there is a discrepancy, a second goal ($[O_2^V]$) is activated to reduce or eliminate the discrepancy, which can be satisfied by (a) changing or devaluing the first goal $[O_1^V]$ (i.e., accommodation), (b) reinterpreting the stimulus (i.e., immunization), or (c) acting in order to change the stimulus (i.e., assimilation).
If assimilation is chosen, a number of action options are compared, and the action option with the highest expected utility is chosen, leading to a third goal to act ($[O_3^V]$, i.e., the intention to act). This third goal also generates a prediction with a certain probability. From this moment on, $[O_1^V]$ not only represents a valued outcome but also a predicted outcome ($[O_1^{v+e}]$). 

After the action is carried out, an outcome is produced. This is the new stimulus, and a representation of this stimulus is fed back as the input to the next run of the cycle. There, it is compared with the representation of the valued and expected outcome $[O_1^{v+e}]$. Any discrepancy with this representation also counts as a reward prediction error that can be used to update the expectancy associated with the chosen action option. 

Affect traces may arise in each phase of the goal-directed cycle. In the discrepancy detection phase, the polarity of affect (positive/negative) is determined by the presence/absence of a discrepancy (this aligns with goal in/congruence in appraisal theories \cite{scherer2001appraisal, moors_demystifying_2022}). The intensity of affect is proportional to the magnitude of the discrepancy and the value of the first goal (this aligns with goal relevance in appraisal theories). In the action selection phase, the polarity of affect (positive/negative) rests on the presence/absence of action options with a sufficiently high expected utility. The intensity of affect is proportional to the expected utility of the to-be-chosen action option. Affect traces may combine to form the total affect derived from the pursuit of the given goal.
For any agent, multiple goals may be active at once. They may be arranged in a hierarchy, where lower-order goals are instrumental for higher-order goals (i.e., vertical complexity). Goals higher up the chain are more abstract (e.g., belonging), whereas goals further down are more concrete (e.g., driving home for Christmas). Furthermore, there may be multiple goal-directed cycles active at the same level of the hierarchy (i.e., horizontal complexity). For instance, an agent may want to have food and to have a drink. Goals that are not in a hierarchical relationship with one another may still influence each other's pursuit through the use of shared and limited resources (e.g., spending money on food or on a drink). The state of goal completion and goal frustration of each goal in the system contributes affect traces from all discrepancy detection and action selection phases from all parallel cycles to form the total affective state of the individual.

\subsection{Existing computational models}
Affective computing has a long-standing tradition of operationalizing psychological models to computationally simulate affective phenomena \cite{broekens2021emotion}.
Early cornerstone architectures \cite{gebhard2005alma, becker2008wasabi} leveraged the OCC model \cite{ortony2022cognitive} as a systematic taxonomy for mapping situational appraisals to affective labels and enabled consistent emotional responses for Socially Intelligent Agents \cite{lugrin2022handbook}. 
However, while these models often incorporate sophisticated temporal dynamics like temporal decay of affect, the generation process remains largely descriptive. 
By relying on predefined, top-down rules (e.g., if goal reached, then joy), these models treat appraisal as a discrete 'snapshot' event rather than a continuous, mechanistic product integrated into an agent’s internal processing cycles.

To bridge this gap, later frameworks \cite{marsella2009ema, scherer2010component, meuleman2013nonlinear, sequeira2014learning, zhang2024simulating} and BDI-based models (Belief-Desire-Intention, \cite{steunebrink2012formal, kaptein2016caaf}) integrated affect into cognitive architectures. 
While these successfully link appraisal to internal causal models and logical resolutions of beliefs, appraisal often remains a specialized evaluative layer that maps cognitive states to affective labels. 
This preserves a functional distinction between the agent’s core cognitive operations and the appraisal process itself, and often struggles to capture the granular, continuous aspects of affect during action execution.
They explain why an emotion occurs based on mental states, but do not necessarily model affect as an emergent property of underlying processing dynamics.
To address the need for greater granularity, another branch of research integrated concepts from Reinforcement Learning (RL) and Predictive Processing to produce models that use reward prediction errors (RPE) (e.g. \cite{broekens_temporal_2018, reisenzein2009emotional}),  homeostatic regulation \cite{lewis2016hedonic, lowe2017predictive} and learning progress \cite{sequeira2011emotion, oudeyer2007intrinsic} as primary drivers of affect. 
These approaches provide continuous signals but remain theoretically underspecified: numerical error signals are often labeled as affective states in ways that lack psychological grounding.

The GDT \cite{moors_demystifying_2022} provides a synthesis that addresses these long-standing tensions by positing that affect is an integral functional product of continuous, multi-level goal-directed cycles, generalizing goal distance as the core driver of affect dynamics.
Unlike descriptive models, it provides a mechanistic explanation, yet unlike purely signal-driven models, it provides psychological and philosophical grounding. 
In this framework, the processes that drive action selection are identical to those computing the affect traces, making affect a direct reflection of the agent's internal dynamics, specifically the interplay between discrepancy detection and action selection. This interplay gives rise to complex affect patterns without the need for dedicated heuristics.

\section{Theoretical Framework}
This work presents a generalized computational architecture of the GDT. Following the psychological theory's lead, we define the core affective processes as abstract functions.
While the current version employs parsimonious default functions to establish a baseline, the framework is explicitly designed to remain agnostic to the specific functional forms. This allows the model to serve as a 'computational laboratory' where different psychological hypotheses about these operations can be tested and compared in future work.
The present computational implementation of the GDT is summarized here.

An agent can have a set of goal states $\boldsymbol{G}$ (Eq. \ref{eq:G}), where each goal state $\boldsymbol{g_m}$ contains target values $t^m_i$ along a given feature dimension $i$ (Eq. \ref{eq:g_m}). We refer to these target values as goals, while goal states are states where goals are fulfilled. The number of features coincides with the number of goals that the agent has. For instance, an agent can have the feature \emph{course-grade} and a particular goal state $\boldsymbol{g_z}$ where that feature equals the maximum grade of that final course exam (e.g., $t_1=10$).
Each goal is also assigned a specific weight value $v_i$ (Eq. \ref{eq:V}), which represents the importance of that specific goal in the current situation.
\allowdisplaybreaks
\begin{align}
        \boldsymbol{G} &= \{g_1, g_2, ... g_M\} \label{eq:G}\\
        \boldsymbol{g_m} &= (t^m_1, t^m_2, ...t^m_I) \label{eq:g_m}\\
        \boldsymbol{v} &= (v_1, v_2, ... v_I) \label{eq:V}
\end{align}

The agent is also in an environment with states $\boldsymbol{S}$ (Eq. \ref{eq:S}) where each state $s_j$ is a vector of scores $f^j_i$ (Eq. \ref{eq:s_j}) along the same features as in Eq. \ref{eq:g_m}. For instance, an agent can be in a state $s_j$ where the feature \emph{course-grade} equals 5.
Further, in each state $s_j$, the agent has access to a set of responses (or actions) $\boldsymbol{R_j}$ (Eq. \ref{eq:R}).
\begin{align}
        \boldsymbol{S} &= \{s_1, s_2, ... s_J\} \label{eq:S}\\
        \boldsymbol{s_j} &= (f^j_1, f^j_2, ...f^j_I) \label{eq:s_j}\\
        \boldsymbol{R_j} &= \{r^j_0, r^j_1, ...r^j_K\} \label{eq:R}
\end{align}

These basic components compose the foundation from which the two affect traces can be calculated, mirroring the discrepancy detection (Eq. \ref{eq:D_i}-\ref{eq:total_A_D}) and action selection phase (Eq. \ref{eq:utility}-\ref{eq:A_R}) and their integration into a total affect (Eq. \ref{eq:A_total_general}-\ref{eq:A_total_simple}).
In the following, index $i$ refers to a feature, $j$ to a state, $m$ to a goal, and $k$ to a response. First, for each state, a discrepancy $D_{ijm}$ is calculated as the distance between target $t^m_i$ and feature $f^j_i$ (Eq. \ref{eq:D_i}). In this paper, we assume $\Delta$ is a simple subtraction ($t^m_i - f^j_i$). However, this is a simplifying assumption that may be further explored in future work.
\begin{equation}
    D_{ijm} = \Delta (t^m_i, f^j_i) \label{eq:D_i}
\end{equation}

The agent further has an expectancy $p^k_{ijm}$ that the discrepancy along goal $t_i$ will be reduced to 0 by taking action $r_k$ (Eq. \ref{eq:expectancy}). 
This expectancy can be further weighted by the certainty parameter $p(s_j)$, which expresses the agent's certainty about the state it is in (Eq. \ref{eq:expectancy}, but also equation \ref{eq:A_D}). For simplicity, we assume parameter $p(s_j)=1$, meaning that the agent is always certain about the state it is in.
Finally, the product of the expectancy and the goal value $v_i$ then yields the expected utility $u^k_{ijm}$ for a given goal and action (Eq. \ref{eq:expUtlity}).

\begin{align}
        p^{k}_{ijm} &= p(D_{ijm} = 0 | s_j, r_k) \cdot p(s_j) \label{eq:expectancy}\\
        u^{k}_{ijm} &= p^{k}_{ijm} \cdot v_i \label{eq:expUtlity}
\end{align}

The GDT states that negative affect (a discrepancy exists, i.e., $D_i>0$) is proportional to the degree of discrepancy from a goal weighted by its value $v_i$. Positive affect (when $D_i = 0$) is proportional to the goal value $v_i$. Both sides can be further weighted by the certainty parameter $p(s_j)$ (Eq. \ref{eq:A_D}).

\begin{align}
        A_{ijm} &= \begin{cases}
            v_i \cdot p(s_j) \quad \text{if} \quad D_{ijm} = 0 | s_j\\
            -v_i \cdot D_{ijm} \cdot p(s_j) \quad \text{if} \quad D_{ijm} > 0 | s_j
        \end{cases} \label{eq:A_D}
\end{align}

The sum of all affect traces arising from the discrepancy detection phase yields $A_D^j$ in state $j$ (Eq. \ref{eq:total_A_D}). Different discrepancies can further be represented by the agent with different representational strengths (similarly to salience) implemented here as the function $h()$. The theory makes no explicit predictions about the form of this function, and for present purposes, we assume $h()$ to be a simple sum.

\begin{align}
        A^j_\mathrm{D} &= h((A_{ijm})_{i,m}) \label{eq:total_A_D}
\end{align}

The affect resulting from the action selection phase is a function of $\boldsymbol{u^k_{ijm}}$. The GDT assumes no specific functional form for this relationship. For the present interpretation, we assume that for one goal:


\begin{equation}
    f(\boldsymbol{u^k_{ijm}})=max(u_{1k}) \ \text{for each action } r_k \in \boldsymbol{R_j}
    \label{eq:utility}
\end{equation}
This calculation yields the total affect derived from action selection, $A_R^j$, in state $j$ (Eq. \ref{eq:A_R}):

\begin{align}
    A^j_\mathrm{R} &= f((u^k_{ijm})_{i,m,k}) \label{eq:A_R}
\end{align}

The total affect $A_{total}^j$ 
at time point $j$ is then a function of (1) the affect derived from discrepancy detection $A_D^j$, (2) the affect derived from action selection $A_R^j$ (Eq. \ref{eq:A_total_general}): 

\begin{align}
    A^j_\mathrm{total} &= a(A^j_\mathrm{D}, A^j_\mathrm{R}) \label{eq:A_total_general}
\end{align}


Again, as the GDT proposes no explicit form for $a()$, we assume a simple interpretation:

\begin{equation}
    A^j_{total} = A^j_{D} + A^j_{R} \label{eq:A_total_simple}
\end{equation}

For the present implementation, we disregard the serial dependence of affect \cite{ariens2023one}, and treat each observation as only depending on the present state of the agent. 
We also start from the case of the agent pursuing only a single goal and receiving only actual (and not anticipated) stimuli.
A further simplifying assumption concerns the time scale of the goal-directed cycles. Goals further down the hierarchy may operate on smaller time-scales than more abstract, higher-order goals (e.g., writing a paragraph in an essay in contrast to getting good grades in school). A situation may also be described on different timescales. For example, the discrepancy-detection and action-selection stages may be separated across different states, or a higher-order goal may be implemented in a less granular state-space than a lower-order goal. For the present purposes, we assume that in each state, a complete goal-directed cycle is run, so that for every state, we have an $A_D^j$ and an $A_R^j$ calculation.

\section{Experimental Methodology}
\subsection{Simulation Framework \& Objectives}
To evaluate the proposed computational model, we implemented an agent within a controlled simulation environment.
While the GDT framework accommodates complex factors, including multiple goals, representational salience, and temporal dependence, this work focuses on the core interaction between stimulus-goal discrepancy and action-selection expectancies. We implement the agent in two tasks:



\begin{enumerate}
    \item A single-step Dice task designed to isolate the affective signatures of both components in their simplest forms 
    \item A multi-step Corridor task that introduces sequential expectancy effects
\end{enumerate}

Both tasks are implemented in multiple versions corresponding to basic and more complex forms of goal progress (binary or gradual) and action expectancies (oblivious or accurate). Following recommendations for open research practices \cite{wessler2021empirical}, all materials, including the computational model, are openly available on OSF\footnote{\url{https://osf.io/bcepk/overview?view_only=b62e62c1867642c3abe130ea3a860b75}}.

\subsection*{Task 1: Dice task}
\begin{figure}[htbp]
\centering
\vspace{-14pt}
\caption{\textit{State transition diagram of the Dice task. V1 denotes the feature scores for the binary task version, while V2 denotes the gradual task version. Arrows represent transitions between states. The arrow pointing from the right back to $s_{throw}$ represents the start of the next trial.}}
\label{fig:states-1}
\vspace{10pt}
\begin{tikzpicture}[
    >=stealth,
    node distance=0.2cm,
    state/.style={circle, draw, minimum size=1cm}
]

\node[state] (sa) {$s_a$};
\node[state, below=of sa] (sb) {$s_b$};
\node[state, below=of sb] (sc) {$s_c$};
\node[state, below=of sc] (sd) {$s_d$};
\node[state, below=of sd] (se) {$s_e$};
\node[state, below=of se] (sf) {$s_f$};

\tikzset{
  f_size/.style={font=\scriptsize}
}

\node[f_size, right=0 cm of sa] (fa1) {$f^a_1=0$};
\node[f_size, right=0 cm of sb] (fb1) {$f^b_1=0$};
\node[f_size, right=0 cm of sc] (fc1) {$f^c_1=0$};
\node[f_size, right=0 cm of sd] (fd1) {$f^d_1=0$};
\node[f_size, right=0 cm of se] (fe1) {$f^e_1=0$};
\node[f_size, right=0 cm of sf] (ff1) {$f^f_1=5$};
\node[above=0cm of fa1] {V1};

\node[f_size, right=0 cm of fa1] (fa2) {$f^a_1=0$};
\node[f_size, right=0 cm of fb1] (fb2) {$f^b_1=1$};
\node[f_size, right=0 cm of fc1] (fc2) {$f^c_1=2$};
\node[f_size, right=0 cm of fd1] (fd2) {$f^d_1=3$};
\node[f_size, right=0 cm of fe1] (fe2) {$f^e_1=4$};
\node[f_size, right=0 cm of ff1] (ff2) {$f^f_1=5$};
\node[above=0cm of fa2] {V2};

\node[state, left=3cm of $(sa)!0.5!(sf)$] (ss) {$s_{start}$};
\node[state, left=1.5cm of $(sa)!0.5!(sf)$] (s_throw) {$s_{throw}$};

\node[below=0 cm of ss] {$f^{start}_1=0$};
\node[above left=0.2 cm and -0.9 cm of s_throw] {$f^{throw}_1=0$};

\draw[->] (ss) -- (s_throw);
\draw[-] (s_throw) -- ($(s_throw)+(1,0)$) coordinate (branchpoint);

\foreach \y in {sa, sb, sc, sd, se, sf}{
    \draw[-] (branchpoint) -- (branchpoint |- \y);
    \draw[->] (branchpoint |- \y) -- (\y);
}

\draw ($(sa.east)+(2.5,0.7)$) coordinate (topline)
      -- ($(sf.east)+(2.5,-0.7)$) coordinate (bottomline);

\path (topline) -- (bottomline) coordinate[midway] (midline);

\coordinate (halfline) at ($(topline)!0.5!(bottomline)$);
\coordinate (rightout) at ($(halfline)+(0.5,0)$);
\coordinate (bottomout) at ($(halfline)+(0.5,-4.3)$);
\coordinate (leftout) at ($(s_throw.south)+(0,-3.65)$);
\coordinate (upout) at ($(ss.west)+(-1,0)$);

\draw[->]
(midline)
-- (rightout)      
-- (bottomout)     
-- (leftout)       
-- (s_throw.south);      

\end{tikzpicture}
\vspace{-5pt}
\end{figure}
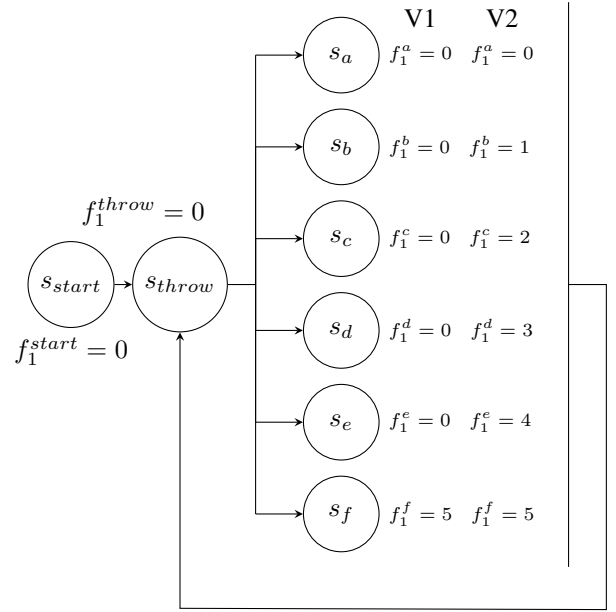

In the Dice task, the agent is throwing a six-sided die. At the start of the task, the agent is in $s_{start}$ with one action option that transitions the agent to $s_{throw}$. In state $s_{throw}$, the agent has one available response, $r_{throw}$, with the transition probability $p(s_{a...f}|s_{throw},r_{throw})=\frac{1}{6}$ to transition to each of six outcome states $s_a, s_b, ..., s_f$ representing the six sides of the die (Figure \ref{fig:states-1}). Each state has a single feature $f^{state}_1$. Goal importance is set to $v_1=1$.

In the task, the agent seeks a specific goal state $g_1=s_f$, where $f^f_1=t_1=5$, representing the simple goal of throwing a 6. In each state, affect is calculated according to formula \ref{eq:A_total_simple}. In state $s_{start}$, the agent detects a discrepancy from its goal state ($D_{1,throw}=5$), but has no goal-conducive actions, leading to negative affect derived purely from discrepancy detection. 
Upon transitioning to $s_{throw}$, the agent is "informed" that it can throw a six-sided die ($r_{throw}$) to potentially reduce its discrepancy with expectancy $p_{1,throw}=1/6$. With the discrepancy remaining the same, this leads to an elevated, albeit still negative total affect combining the affect traces from the discrepancy detection and action selection stages. After throwing, the agent ends up in a terminal state where affect is again purely driven by the discrepancy detection stage.

Two versions of this task were implemented (Binary vs. Gradual) to contrast "hit-or-miss" goals with goals where state features reflect proximity to the target. In the Binary version (V1 in Figure \ref{fig:states-1}), the feature scores in each state are set up so that only $s_f$ results in no discrepancy (positive affect), while all other states result in a discrepancy of -5 (negative affect). In the Gradual version (V2 in Figure \ref{fig:states-1}), the feature scores $f^{state}_1 \in \{0, 1, 2, 3, 4, 5\}$ reflect proximity to the goal of throwing a 6 ($s_f$). This introduces an additional gradient to the $A_D$ values in terminal states mirroring a human reacting differently when throwing a 5 than when throwing a 1 based on a naive understanding of probability. 
The Dice task presents a way to disentangle the two affective components and their contribution to the affect calculation. In $s_{start}$, we see affect derived only from the discrepancy when $D_{1j}>0$. In $s_{throw}$, we see the basic interplay of discrepancy and expectancy in affect. Finally, in $s_f$, we see affect only from the discrepancy detection when $D_{1j}=0$ (i.e., the goal is reached).

\subsection*{Task 2: Corridor task}
\begin{figure}[htbp]
\centering
\vspace{-14pt}
\caption{\textit{State transition diagram for the Corridor task (binary version). Arrows represent transitions between states.}}
\label{fig:states-3}
\vspace{10pt}
\begin{tikzpicture}[
    >=stealth,
    node distance=0.3cm,
    state/.style={circle, draw, minimum size=1cm}
]

\node[state] (ss) {$s_{start}$};
\node[state, right=of ss] (sa) {$s_a$};
\node[state, right=of sa] (sb) {$s_b$};
\node[state, right=of sb] (sc) {$s_c$};
\node[state, right=of sc] (sd) {$s_d$};
\node[state, right=of sd] (se) {$s_e$};

\coordinate (midpoint) at ($(ss)!0.5!(se)$);
\pgfmathsetmacro{\trapoffset}{0.8};
\node[state, above = \trapoffset cm of midpoint] (strap) {$s_{trap}$};

\node[left=0 cm of strap, yshift=2 mm] (fstrap1) {V1 $f^{trap}_1=0$};
\node[right=0 cm of strap, yshift=2 mm] (fstrap2) {V2 $f^{trap}_1=-3$};
\node[below=0 cm of ss] (fs1) {$f^{start}_1=0$};
\node[below=0 cm of sa] (fa1) {$f^a_1=0$};
\node[below=0 cm of sb] (fb1) {$f^b_1=0$};
\node[below=0 cm of sc] (fc1) {$f^c_1=0$};
\node[below=0 cm of sd] (fd1) {$f^d_1=0$};
\node[below=0 cm of se] (fe1) {$f^e_1=5$};
\node[left=0 cm of fs1] (f1) {V1};

\node[below=0 cm of fs1] (fs2) {$f^{start}_1=0$};
\node[below=0 cm of fa1] (fa2) {$f^a_1=1$};
\node[below=0 cm of fb1] (fb2) {$f^b_1=2$};
\node[below=0 cm of fc1] (fc2) {$f^c_1=3$};
\node[below=0 cm of fd1] (fd2) {$f^d_1=4$};
\node[below=0 cm of fe1] (fe2) {$f^e_1=5$};
\node[left=0 cm of fs2] (f2) {V2};

\foreach \x/\y in {ss/sa, sa/sb, sb/sc, sc/sd, sd/se} {
    \draw[->] (\x) -- (\y);
    \coordinate (temp_mid) at ($(\x)!0.5!(\y)$);
    \ifthenelse{\equal{\x}{sb}}{
        \draw[->] (temp_mid) -- (strap);
    }{
        \coordinate (temp_up) at ($(temp_mid)+(0, \trapoffset)$);
        \draw[-] (temp_mid) -- (temp_up);
        \draw[->] (temp_up) -- (strap);
    }
}
\end{tikzpicture}
\vspace{-30pt}
\end{figure}
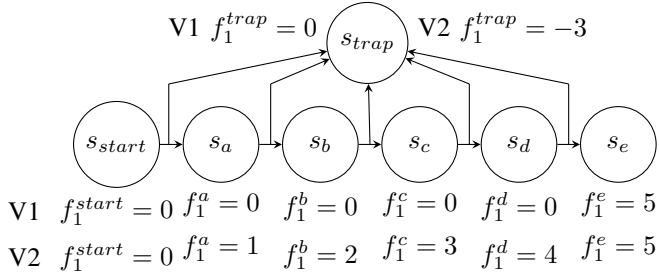

In the Corridor task, the agent walks along a corridor, starting at the leftmost position ($s_{start}$ in Fig. \ref{fig:states-3}). In each state, the agent has a single behavior option, $r_1$, that transitions the agent (a) to the next state to the right with a transition probability of 0.9 or (b) into a trap, $s_{trap}$, with a transition probability of 0.1. Each state has a single feature $f^{state}_1$. 
In the trap state, $f^{trap}_1$ is either 0, for not reaching the goal in the Binary version, or -3 in the Gradual version, differentiating failure from the starting state. 
Goal value is again set to $v_1=1$. The agent's goal $g_1$ is to reach the rightmost end of the corridor, $g_1 = s_e$ where $f^e_1=t_1=5$. Again, at each state, affect is calculated according to Eq. \ref{eq:A_total_simple}. 

The Corridor task is sequential, unlike the Dice task. To account for multiple steps in the GDT framework, we could posit lower-order goals that implement the higher-order goal $g_1$ step-by-step. Another approach (which may or may not be equivalent) is to calculate the expectancy $p^{step}_{1j1}$ by discounting the subjective probability of transitions across the steps required for the agent to reach its goal state $g_1$. For this latter approach, we define a further element of our model, $p'_{x|y}$, the subjective probabilities of transitioning from state $x$ to state $y$. For an agent with a single goal in state $s_j$, with $n_{steps}$ being the number of steps required to reach $g_1$:
\begin{equation}
    p^{step}_{1j1}=p(D_1=0|s_j,r_1)=\prod_{i=j}^{n_{steps}}(p'_{i+1|i})
    \end{equation}
If all $p'_{x|y}$ are equal, this simplifies to:
\begin{equation}
    p^{step}_{1j1} = (p')^{n_{steps}} \ \ \text{where all} \ p'_{x|y} =p'
\end{equation}


An oblivious agent assumes perfect subjective transition probabilities $p'=1$, while an agent knowing about the trap door accurately discounts expectancies aligned with true transition probabilities ($p(s'|s,r)=0.9$ and $p(trap|s,r)=0.1$): 

{
\renewcommand{\arraystretch}{1.3}
\begin{table}[ht]
    \centering
    \begin{tabular}{cccccc}
         & $s_{start}$ & $s_a$ & $s_b$ & $s_c$ & $s_d$ \\
         \midrule
        $p^{step}_{1j1}$ & 0.59049 & 0.6561 & 0.729 & 0.81 & 0.9
    \end{tabular}
\end{table}
}
Similarly to the Dice task, four versions of the Corridor task were set up along two axes: (1) a Binary and a Gradual version (equal to the dice task, V1/V2 in Fig. \ref{fig:states-3}), and (2) an Oblivious and an Accurate version that contrasts agents with subjective transition probabilities of 1, against those whose subjective probabilities align with true transition probabilities. 

Taken together, the binary discrepancy gives a flat affect component in $A_D$, while the gradual discrepancy gives a linearly increasing affect component in $A_D$. In parallel, expectancies give a flat affect component in $A_R$ for the oblivious agent, and a non-linearly increasing 
one for the accurate agent. These four combinations present four distinct affect profiles produced by the interaction of the goal discrepancies and the agent's expectancies according to the GDT.

\section{Results}

\begin{figure}[ht]
    \centering
    \includegraphics[width=0.9\linewidth, frame]{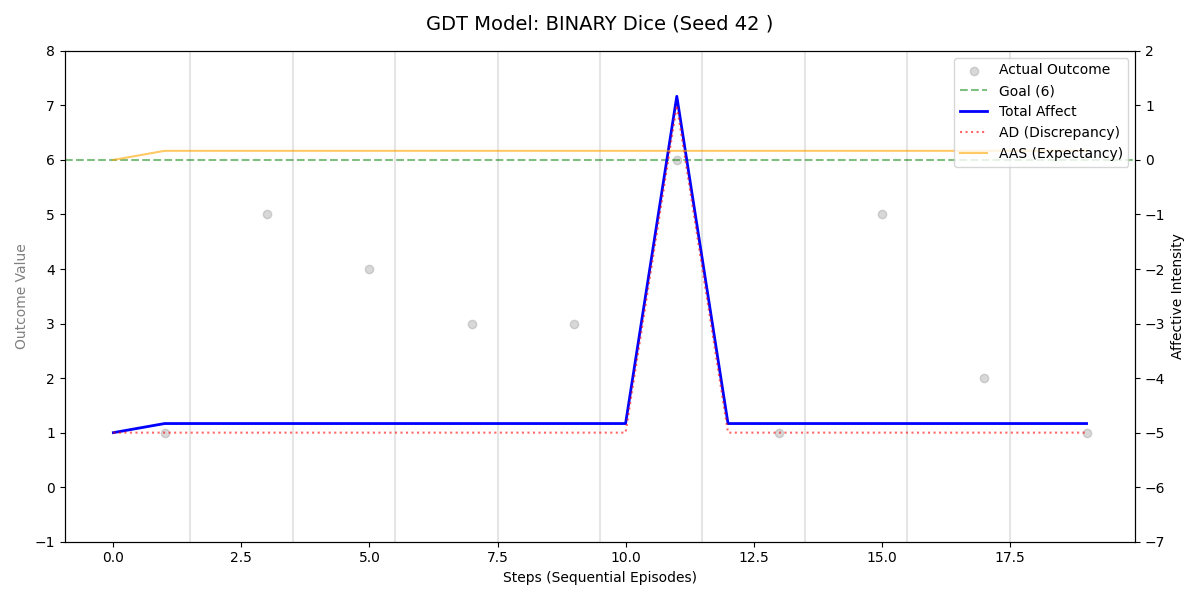}
    \caption{Affective response in the Binary Dice Task ($t_1=5.0$). $A_D$ remains stagnant at a baseline of $-5.0$ for all non-target throws, $A_R$ is stable at $1/6$ with exception of $s_{start}$. Total Affect is elevated from $A_D$ by the size of $A_R$. Spikes in Total Affect occur only when the specific goal state is achieved.}
    \label{fig:dice_bin}
    \vspace{-10pt}
\end{figure}

\begin{figure}[ht]
    \centering
    \includegraphics[width=0.9\linewidth, frame]{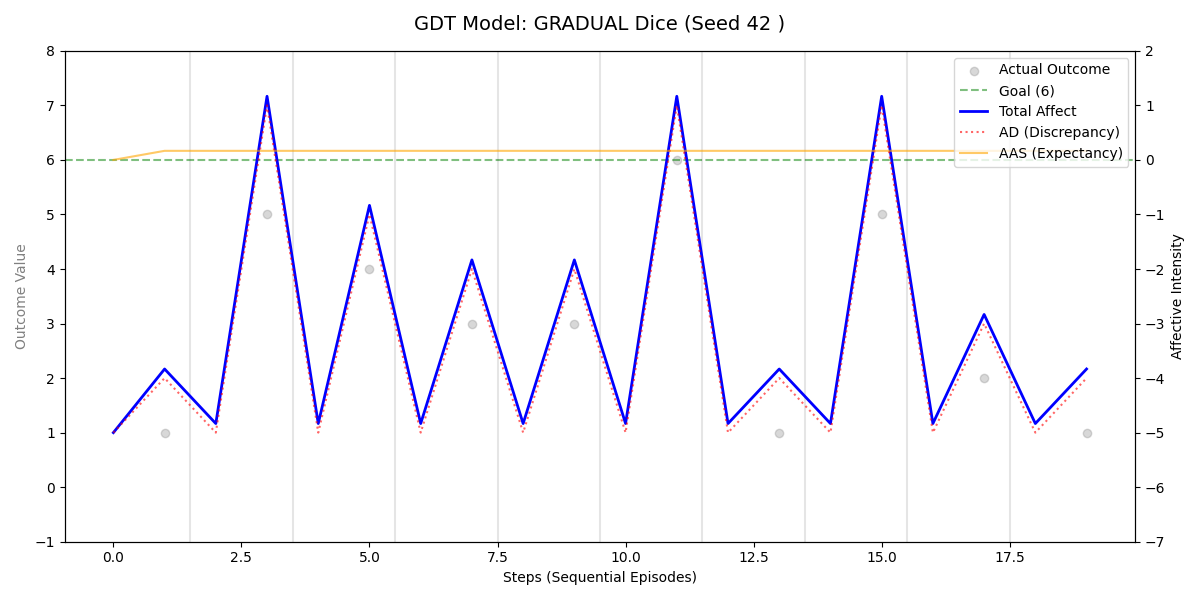}
    \caption{Gradual Dice Task ($t_1=5.0$). $A_D$ fluctuates proportionally to the numerical distance between the outcome value and the target. Near-miss outcomes result in higher intermediate affect compared to distant rolls.}
    \label{fig:dice_grad}
    \vspace{-16pt}
\end{figure}


In the Dice task, we compared binary and gradual discrepancies. 
In the binary version (Fig. \ref{fig:dice_bin}), affect intensity remained at a negative baseline for all outcomes except the goal ($s_f$). Upon achieving the goal, a sharp positive spike in total affect was observed, driven entirely by the elimination of discrepancy ($D_{ij}=0$).
There was a positive shift in total affect when the agent transitioned from a state with no viable behavior options ($s_{start}$) to one with viable responses ($s_{throw}$).
In the gradual mode, total affect scaled proportionally to the face value of the dice. 
Higher rolls resulted in higher affective intensity due to the reduction in distance to the target value (Fig. \ref{fig:dice_grad}). 
\begin{figure}[ht]
    \centering
    \vspace{-2pt}
    \includegraphics[width=0.9\linewidth, frame]{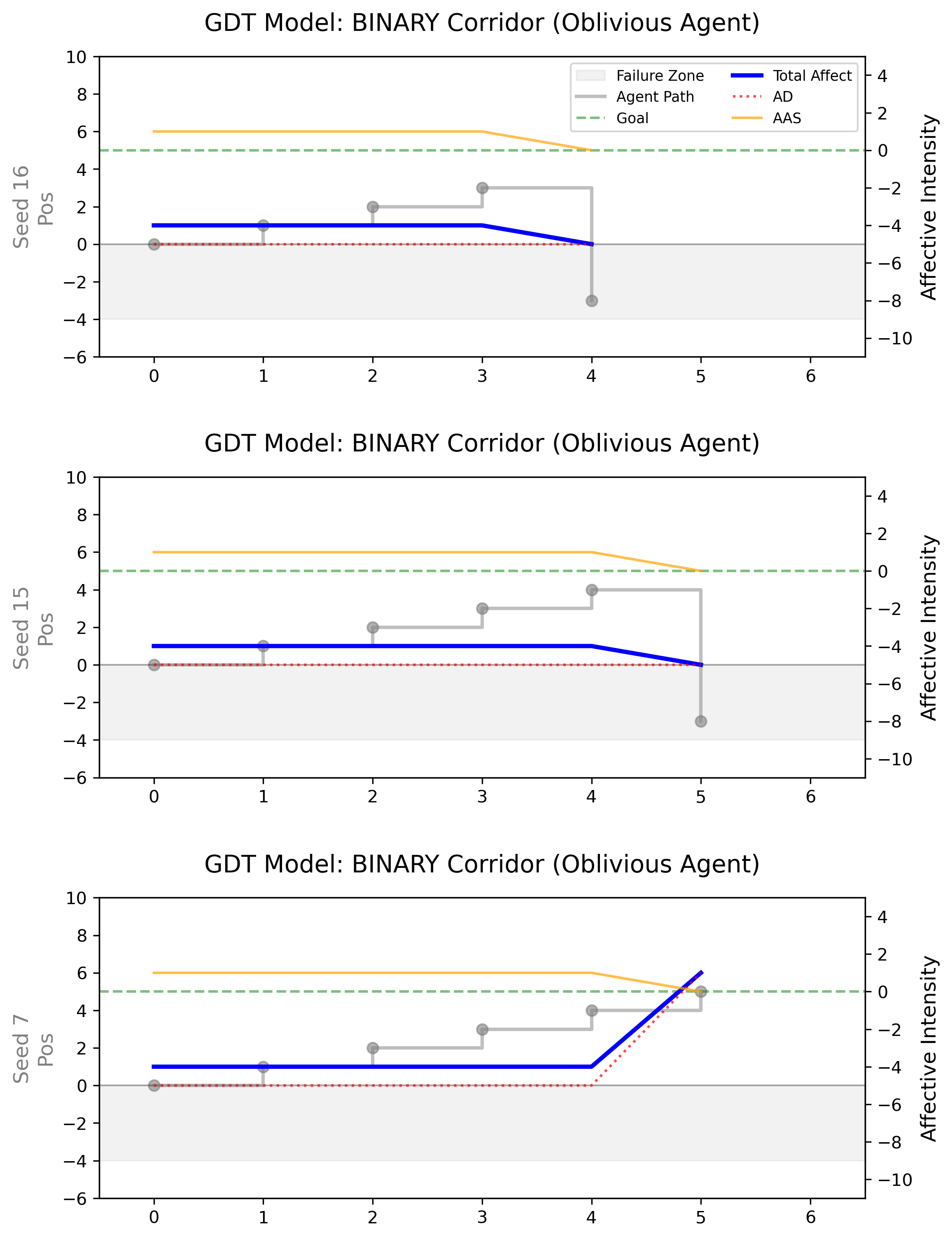}
    \caption{Binary Corridor Task (Oblivious): Affect remains at a flat negative baseline because the agent perceives neither proximity to the goal ($A_{D}$) nor changes in trap risk ($A_{R}$). The top panel shows an agent falling into the trap at step 4, the middle panel shows falling into the trap at step 5, while the bottom panel shows an agent reaching the goal state.}
    \label{fig:corridor_bin_ob_stack}
    \vspace{-14pt}
\end{figure}

\begin{figure}[ht]
    \centering
    \includegraphics[width=0.9\linewidth, frame]{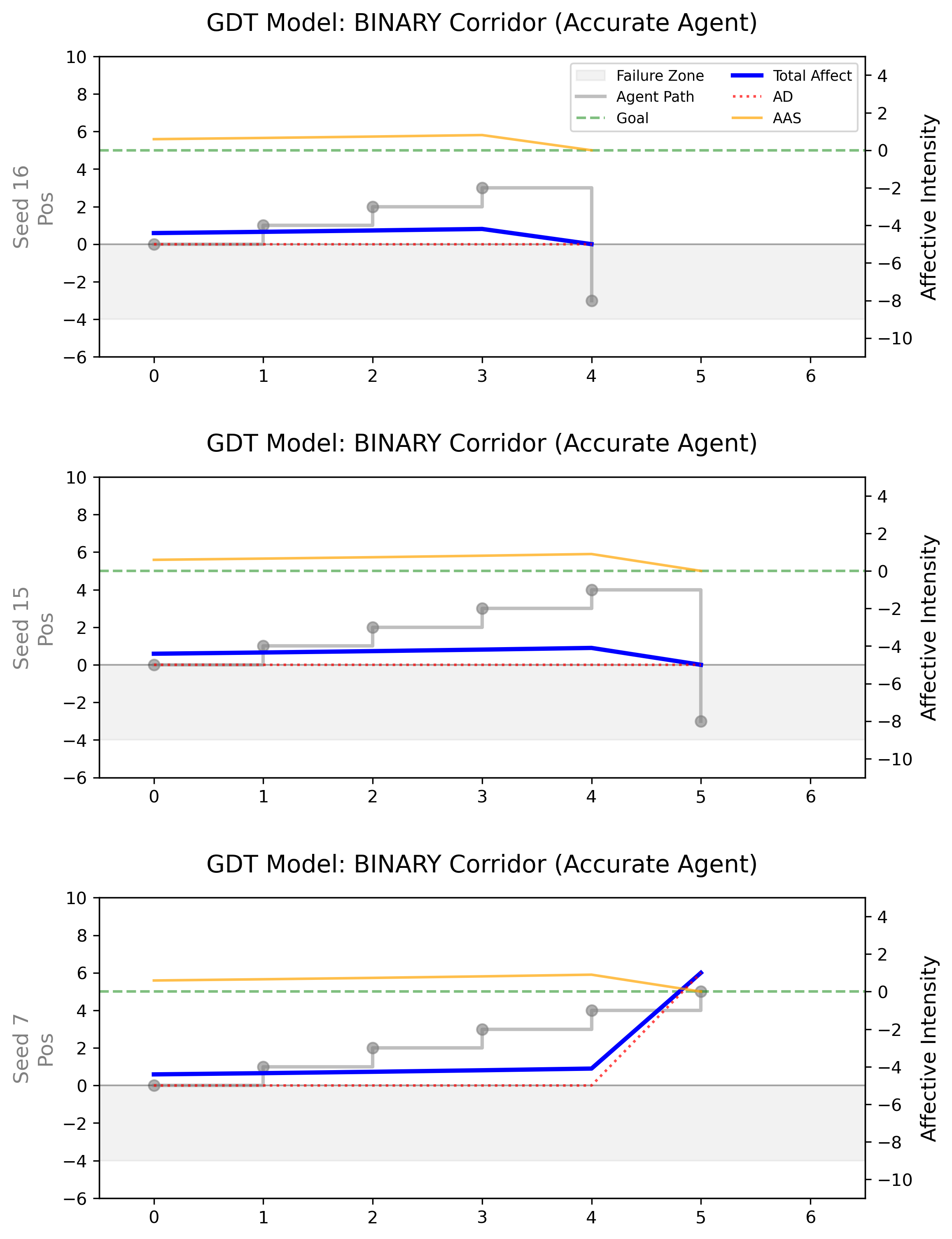}
    \caption{Binary Corridor Task (Accurate): A subtle upward trend in total affect emerges from the non-linear "expectancy lift" ($A_{R}$) as the agent moves further from the start and away from cumulative trap risk. The top panel shows an agent falling into the trap at step 4, the middle panel shows falling into the trap at step 5, while the bottom panel shows an agent reaching the goal state.}
    \label{fig:corridor_bin_acc_stack}
    \vspace{-20pt}
\end{figure}

The Corridor task results illustrate the longitudinal development of affect during multi-step goal pursuit. 
Two versions of discrepancy detection (Binary vs. Gradual) and agent expectancies (Oblivious vs. Accurate) were examined.
\begin{figure}[ht]
    \centering
    \includegraphics[width=0.9\linewidth, frame]{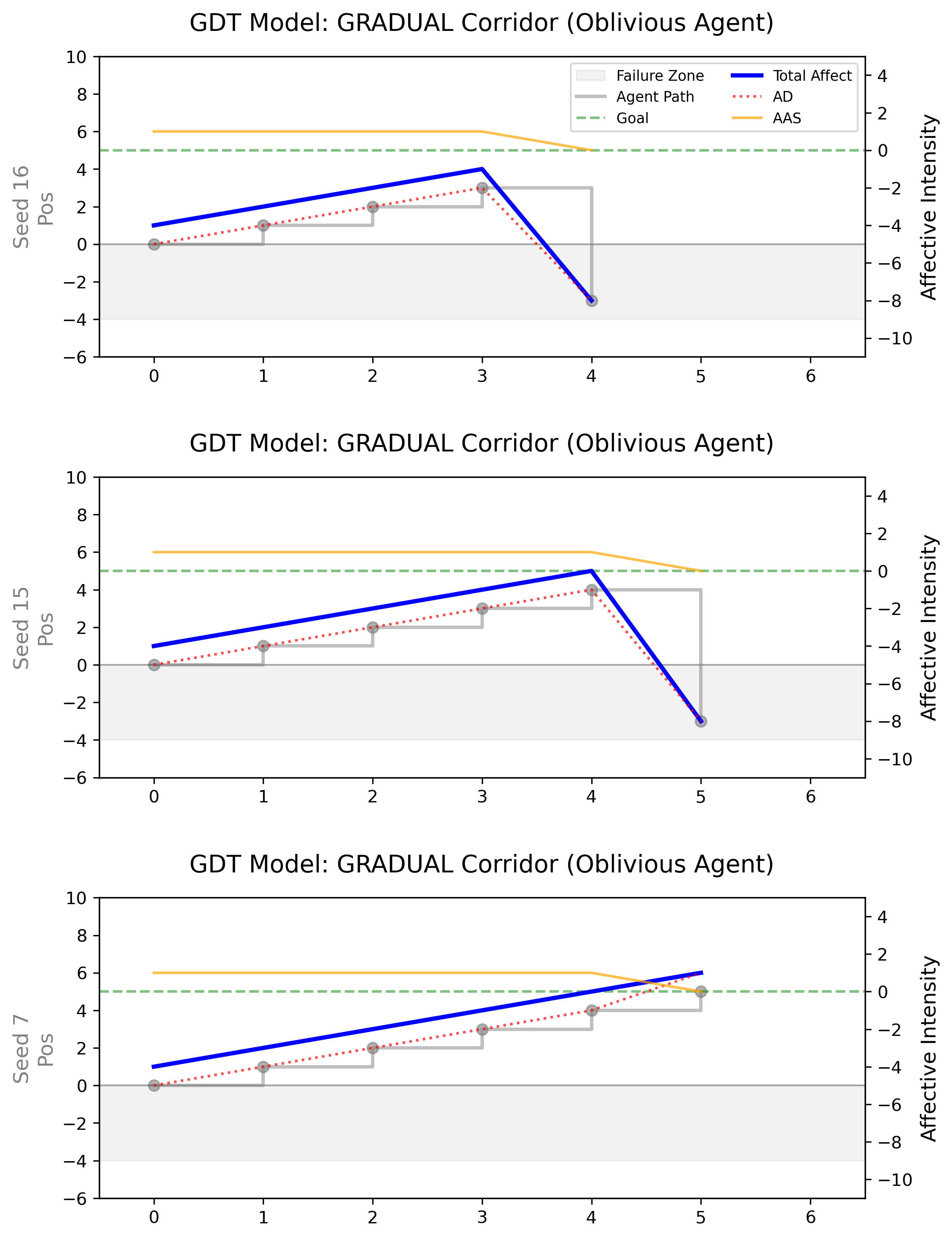}
    \caption{Gradual Corridor Task (Oblivious): Total affect increases linearly, driven by the steady reduction in numerical goal discrepancy ($A_{D}$) as the agent physically approaches the target. The top panel shows an agent falling into the trap at step 4, the middle panel shows falling into the trap at step 5, while the bottom panel shows an agent reaching the goal state.}
    \label{fig:corridor_grad_ob_stack}
    \vspace{-14pt}
\end{figure}
In the binary-oblivious condition,
both the $A_D$ and $A_R$ components remained entirely horizontal (Fig. \ref{fig:corridor_bin_ob_stack}). Because the agent did not perceive "closeness" to the goal and maintained a constant expectancy, the total affective intensity remained at a negative baseline until the final step (goal state $g = s_e$ or trap state $s_{trap}$).
\begin{figure}[ht]
    \centering
    \includegraphics[width=0.9\linewidth, frame]{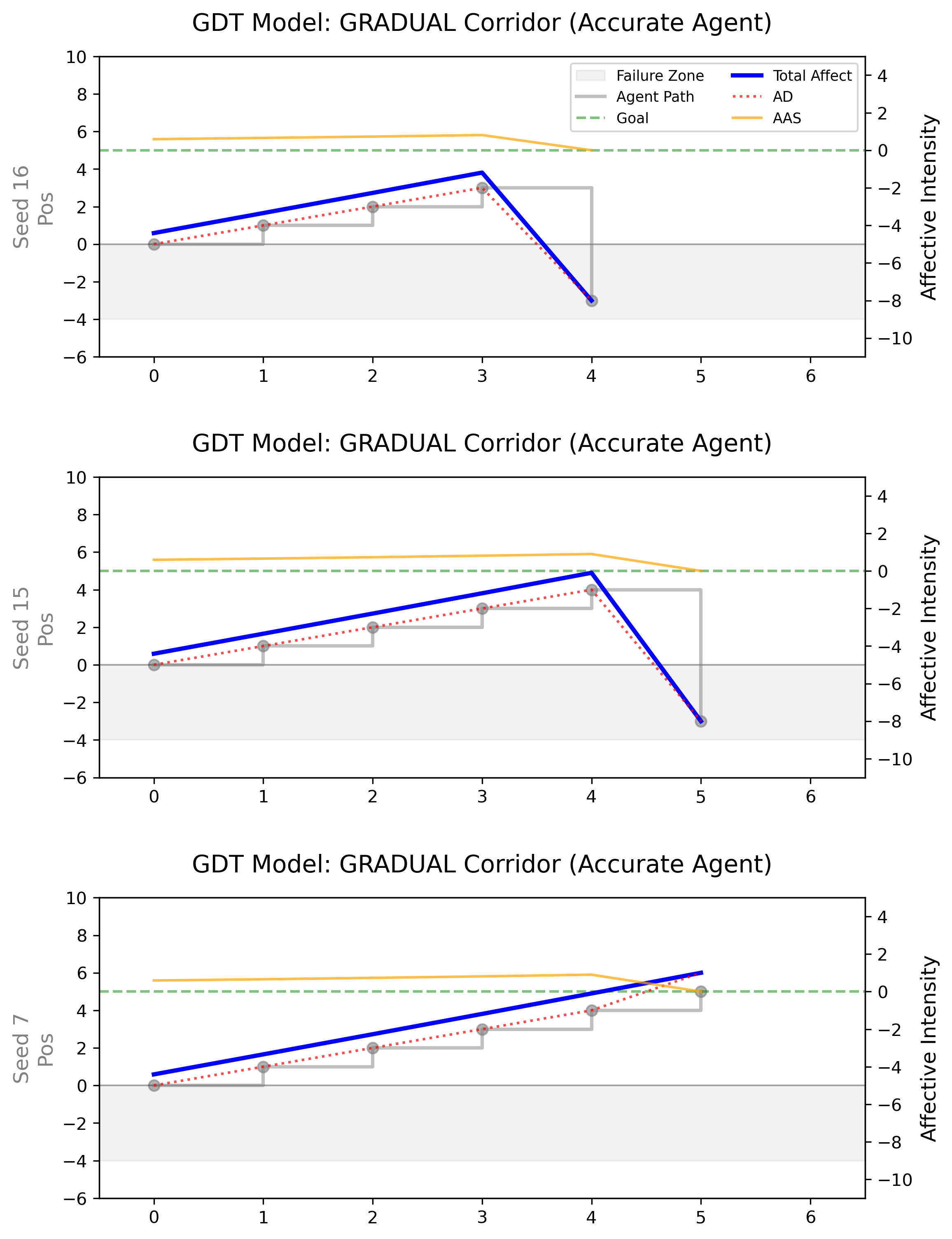}
    \caption{Gradual Corridor Task (Accurate): The linear reduction in discrepancy ($A_{D}$) and the non-linear expectancy lift ($A_{R}$) combine to produce peak affective intensity prior to reaching the goal. The top panel shows an agent falling into the trap at step 5, the middle panel shows falling into the trap at step 6, while the bottom panel shows an agent reaching the goal state.}
    \label{fig:corridor_grad_acc_stack}
    \vspace{-14pt}
\end{figure}
When accurate expectancy was introduced to the binary version, a subtle upward trend in total affect emerged (Fig. \ref{fig:corridor_bin_acc_stack}). While the $A_D$ remained flat (as no progress was perceived), the $A_R$ component increased as the agent moved further from the start, reflecting the decreased cumulative probability of hitting a trap. 
In the gradual oblivious mode, the agent’s affect was driven primarily by decreasing discrepancies (Fig. \ref{fig:corridor_grad_ob_stack}). A linear increase in total affect was observed as the agent approached the goal, characterized by a steady upward slope in the $A_D$ component. The $A_R$ component remained static as the expectancies were the same in each state.
The gradual-accurate condition represented the most complex affective profile (Fig. \ref{fig:corridor_grad_acc_stack}). Total affect was modulated by both the linear reduction in discrepancy ($A_D$) and the non-linear "expectancy lift" ($A_R$) as the agent neared success, resulting in the highest affective intensity prior to reaching the goal.

Across all versions, reaching the goal produced a peak in intensity where $D_{ij}=0$ while $A_R=1$.
Failure in the gradual versions produced a "crash": $A_D$ plummeted to a failure baseline (feature value -3.0) and $A_R$ dropped to zero.
In the binary versions, the $A_D$ component did not drop further upon hitting the trap; it remained at the same negative baseline as the rest of the corridor. However, a drop in total affect was still recorded due to the $A_R$ component plummeting to zero, representing the loss of all remaining expectancy of success.

\section{Discussion}
By systematically examining the individual contributions of each component and their dynamics, we 
illustrate in computo the core GDT assumption that total affect is a predictable, joint outcome of goal discrepancy and expected utility.
A critical feature of the computational model demonstrated in both tasks is that non-zero expectancies provide a "lift" to total affect even when the actual discrepancy remains negative. In the Dice task, states with a viable behavior option create an anticipatory "lift". 
The Corridor task further shows this 
in more complex affect profiles arising from the same underlying equations applied to a more complex situation. In the binary-oblivious condition, the lack of granularity in discrepancies and the constant expected utility of behavior options resulted in flat, stagnant affect (as in the Dice task). Gradual discrepancies introduced a linearly increasing element to total affect, while compounded expectancies added a non-linearly increasing affect trace. 
 Since, in the current implementation of the discrepancy calculation, $A_D$ was not scaled, but calculated by a simple difference, total affect was mainly driven by  $A_D$ with a smaller contribution from $A_R$ (i.e., expected utility ranges from 0 to 1, discrepancy from 0 to 5). 
Following the expectancy lift, even when there is no feedback on goal progress (Binary version), a drop in total affect is recorded in $s_{trap}$ solely due to the loss of viable behavior options ($A_R=0$). In the Gradual version, failure manifested as a simultaneous "crash" of both components: $A_D$ plummeting to a failure baseline in $s_{trap}$ and $A_R$ dropping to zero. This illustrates how the GDT computational framework provides a mechanistic, computational explanation for positive and negative relative anticipatory affect, showing it as arising naturally from action selection without requiring a separate "anticipation" module, while also differentiating loss ($A_D$ drop) from disappointment ($A_R$ drop).


\subsection{Limitations and Implications}
Several limitations provide a roadmap for future work. First, in the current version of the framework, several simplifying assumptions were made regarding the specific forms of the functions involved, due to the lack of specific predictions made by the GDT. In the future, joint psychological and affective computing research needs to iteratively test different functional forms against empirical data to further refine those functions. 
Second, the current model lacks learning dynamics and transition probabilities are static. Integrating reinforcement learning in future experiments would allow us to observe how affect evolves as an agent’s mastery of the environment increases. Third, we have omitted temporal decay.
In biological systems, affective states persist and color subsequent appraisals; adding a "leakage" or momentum parameter to the discrepancy calculations would increase ecological validity. Finally,  the model must be tested against human empirical data 
to validate that the affect profiles produced by the model correlate with human physiological or self-report trajectories.

By formalizing the 
GDT in a scalable way, this work provides a framework where affective states are a functional byproduct of behavior rather than just descriptive labels.
Therefore, it moves beyond "black-box" affect models towards granular, mechanistic theories that can be fully simulated and inspected.
Crucially, this architecture serves as a 'computational toolbox', allowing future research to refine specific formulas without rewriting the underlying framework.
This facilitates experimental paradigms based on a "simulation-empiry" loop: allowing researchers to predict how experimental manipulations will manifest in affective profiles before running empirical studies with human participants. 
In line with this, the simplicity of the tasks (Dice, Corridor) is an intentional design choice to ensure results are comparable to human experimental paradigms, facilitating empirical validation.

\section{Conclusion}
In this paper, the first high-fidelity computational implementation of the Goal-Directed Theory (GDT) of affect was presented. By formalizing the psychological mechanisms of the GDT into a simulatable framework, we systematically disentangle the contributions of goal discrepancy and action selection to affective experience. 
Additionally, we demonstrate that complex affective profiles such as the gradual "lift" of anticipation, can emerge naturally as functional byproducts of active behavior, without requiring separate, dedicated modules.
This architecture serves as a "computational toolbox" for affective science for evaluating the GDT and develop it further.
By shifting the focus from "black-box" heuristics to a transparent, granular architecture, this model bridges the gap between high-level cognitive appraisals and low-level active processing. As every component in the formalization maps directly to states that can be controlled and measured in human subjects, this work paves the way for a continuous "simulation-empiry" research loop. 

\section*{Ethical Impact Statement}
 In accordance with the ACII 2026 submission guidelines and ethical standards, we identify the following ethical considerations and potential impacts:

\subsection*{Direct Impact and Research Integrity}
The primary impact of this work is foundational and scientific. By providing a transparent ``computational laboratory'' to test psychological hypotheses, this research bridges the gap between cognitive appraisals and mechanistic behavior. The methodology promotes research integrity by enabling a ``simulation-empiry'' loop, allowing for the prediction of human affective responses under strictly controlled, reproducible conditions.

\subsection*{Data and Simulation Ethics}
The current study relies entirely on principled computational simulations. It does not involve human participants, animal subjects, or the collection of sensitive personal data. Consequently, no Institutional Review Board (IRB) approval was required for this stage of the research. To ensure transparency and prevent scientific misinformation, all materials, including simulation code and task parameters, are documented and hosted on the Open Science Framework (OSF).

\subsection*{Societal Implications and Potential Misuse}
\begin{itemize}
    \item \textbf{Socially Intelligent Agents:} This model can be used to develop agents with more nuanced, psychologically grounded affective profiles. While this enhances human-computer interaction, we acknowledge the risk that such systems could be used to manipulate user emotions. We advocate for the use of this model in ways that prioritize user autonomy and emotional well-being.
    \item \textbf{Interpretability vs. Manipulation:} Unlike ``black-box'' affective models, our implementation focuses on granular, mechanistic transparency. This design choice is intended to mitigate the risks of opaque AI by making the underlying triggers of agent ``affect'' auditable.
    \item \textbf{Clinical Considerations:} While this model provides insights into goal frustration and discrepancy detection—concepts relevant to mental health—it is a foundational theory and not a clinical tool. We explicitly state that this model should not be used for medical diagnosis or treatment without extensive empirical validation and multidisciplinary ethical oversight.
\end{itemize}

\subsection*{Mitigation Strategies}
By mapping computational components directly to measurable human states, we provide a framework that is inherently more explainable than heuristic-based models. Future research intended to validate these computational signatures against human empirical data will be conducted under strict ethical guidelines, ensuring informed consent and the protection of participant privacy.

\section*{Acknowledgment}
This research is partly sponsored by the Hybrid Intelligence project, grant number 024.004.022. The work of Tamás Szűcs and Agnes Moors is supported by the Research Fund of KU Leuven (grant number C14/23/062). Special thanks to Noah Andries and Felix Kleuker.


\bibliographystyle{IEEEtran}
\bibliography{CR_references}

\end{document}